\UseRawInputEncoding
\documentclass{article}
\pdfoutput=1

\usepackage{times}
\usepackage{epsfig}
\usepackage{graphicx}
\usepackage{amsmath}
\usepackage{amssymb}

\usepackage{booktabs}
\usepackage{dsfont}

\usepackage{setspace}
\usepackage{bm}
\usepackage{multirow}
\usepackage{pifont}
\usepackage{balance}
\usepackage{makecell}
\usepackage{colortbl}
\usepackage{cuted} 

\usepackage{multicol}
\usepackage{adjustbox}
\usepackage{wrapfig}

\usepackage[preprint]{style}

\usepackage[utf8]{inputenc} 
\usepackage[T1]{fontenc}    
\usepackage{hyperref}       
\usepackage{url}            
\usepackage{booktabs}       
\usepackage{amsfonts}       
\usepackage{nicefrac}       
\usepackage{microtype}      
\usepackage{xcolor}         

\title{SAM Fails to Segment Anything? -- SAM-Adapter: Adapting SAM in Underperformed Scenes: Camouflage, Shadow, Medical Image Segmentation, and More}

\author{Tianrun Chen$^{1,2 +*}$\and Lanyun Zhu$^{4 +}$\and Chaotao Ding$^{3 +}$\and Runlong Cao$^{3 +}$ \\ \and Yan Wang $^{6}$ \and Zejian Li $^{5}$ \and Lingyun Sun $^{2}$ \and Papa Mao$^{1}$ \\\and Ying Zang$^{3*}$\\
\\
\small First Online: 14 April, 2023
\\
\small $^+$ Equal Contribution   $^*$ Corresponding Author \\$\{$tianrun.chen@zju.edu.cn; 02750@zjhu.edu.cn$\}$
\\
\small $^{1}$KOKONI, Moxin (Huzhou) Tech. Co., LTD, Huzhou, Zhejiang, P.R. China.\\ 
\small $^{2}$College of Computer Science and Technology, Zhejiang University, Hangzhou, Zhejiang, P.R. China.\\ 
\small $^{3}$School of Information Engineering, Huzhou University, Huzhou, Zhejiang, P.R. China.\\ 
\small $^{4}$ Information Systems Technology
and Design Pillar, Singapore University of Technology and Design, Singapore.\\
\small $^{5}$ School of Software Technology, Zhejiang University, Hangzhou, Zhejiang, P.R. China.\\ 
\small $^{6}$ School of Instrumentation and Optoelectronic Engineering, Beihang University, Beijing, P.R. China.\\ 
\\
Project Page: \href{http://tianrun-chen.github.io/SAM-Adaptor/}{http://tianrun-chen.github.io/SAM-Adaptor/}
}
\begin{document}

\maketitle

\begin{abstract}
The emergence of large models, also known as foundation models, has brought significant advancements to AI research. One such model is Segment Anything (SAM), which is designed for image segmentation tasks. However, as with other foundation models, our experimental findings suggest that SAM may fail or perform poorly in certain segmentation tasks, such as shadow detection and camouflaged object detection (concealed object detection). This study first paves the way for applying the large pre-trained image segmentation model SAM to these downstream tasks, even in situations where SAM performs poorly. Rather than fine-tuning the SAM network, we propose \textbf{SAM-Adapter}, which incorporates domain-specific information or visual prompts into the segmentation network by using simple yet effective adapters. By integrating task-specific knowledge with general knowledge learnt by the large model, SAM-Adapter can significantly elevate the performance of SAM in challenging tasks as shown in extensive experiments. We can even outperform task-specific network models and achieve state-of-the-art performance in the task we tested: camouflaged object detection, shadow detection. We also tested polyp segmentation (medical image segmentation) and achieves better results. We believe our work opens up opportunities for utilizing SAM in downstream tasks, with potential applications in various fields, including medical image processing, agriculture, remote sensing, and more. 
  
\end{abstract}

\section{Introduction}
 AI research has witnessed a paradigm shift with models trained on vast amounts of data at scale. These models, or known as foundation models, such as BERT, DALL-E, and GPT-3 have shown promising results in many language or vision tasks\cite{bommasani2021opportunities}.  Recently, among the foundation models, Segment Anything (SAM)\cite{SAM} has a distinct position as a generic image segmentation model trained on the large visual corpus \cite{SAM}. It has been demonstrated that SAM has successful segmentation capabilities in diverse scenarios, which makes it a groundbreaking step toward image segmentation and related fields of computer vision.
 
  However, as computer vision encompasses a broad spectrum of problems, SAM's incompleteness is evident, which is similar to other foundation models since the training data cannot encompass the entire corpus, and working scenarios are subject to variation \cite{bommasani2021opportunities}. In this study, we first test SAM in some challenging low-level structural segmentation tasks including camouflaged object detection (concealed scenes) and shadow detection, and we find that the SAM model trained on general images cannot perfectly "Segment Anything" in these cases. 
 
 As such, a crucial research problem is: \textit{How to harness the capabilities acquired by large models from massive corpora and leverage them to benefit downstream tasks?}
 
 Here, we introduce the \textbf{SAM-Adapter}, which serves as a solution to the research problem mentioned above. \textit{\textbf{This pioneering work is the first attempt to adapt the large pre-trained image segmentation model SAM to specific downstream tasks with enhanced performance.}} As its name states, \textbf{SAM-Adapter} is a very simple yet effective adaptation technique that leverages internal knowledge and external control signal. Specifically, it is a lightweight model that can learn alignment with a relatively small amount of data and serves as an additional network to inject task-specific guidance information from the samples of that task. Information is conveyed to the network using visual prompts \cite{yang2023exploring, liu2023explicit}, which has been demonstrated to be efficient and effective in adapting a frozen large foundation model to many downstream tasks with a minimum number of additional trainable parameters.
 
 Specifically, we show that our method is:
 \begin{itemize}
       \item \textbf{Generalizable}: SAM-Adapter can be directly applied to customized datasets of various tasks to enhance performance with the assistance of SAM. 
       \item \textbf{Composable}: It is effortless to combine multiple explicit conditions to fine-tune SAM with multi-condition control.
\end{itemize}

We perform extensive experiments on multiple tasks and datasets, including ISTD for shadow detection \cite{wang2018stacked} and COD10K \cite{fan2020camouflaged}, CHAMELEON \cite{skurowski2018animal}, CAMO \cite{le2019anabranch} for camouflaged object detection task, and kvasir-SEG \cite{jha2020kvasir} for polyp segmentation (medical image segmentation) task. Benefiting from the capability of SAM and our SAM-Adapter, our method achieves state-of-the-art (SOTA) performance on both tasks. The contributions of this work can be summarized as follows:
\begin{itemize}
    \item First, we pioneer the analysis of the incompleteness of the Segment Anything (SAM) model as a foundation model and propose a research problem of how to utilize the SAM model to serve downstream tasks. 
    \item Second, we are the first to propose the adaptation approach, \textbf{SAM-Adapter}, to adapt SAM to downstream tasks and achieve enhanced performance. The adapter integrates the task-specific knowledge with general knowledge learnt by the large model. The task-specific knowledge can be flexibly designed. 
    \item Third, despite SAM's backbone being a simple plain model lacking specialized structures tailored for the two specific downstream tasks, our approach still surpasses existing methods and attains state-of-the-art (SOTA) performance in these downstream tasks.
\end{itemize}

To the best of our knowledge, this work pioneers to demonstrate the exceptional ability of SAM to transfer to other specific data domains with remarkable accuracy. While we only tested it on a few datasets, we expect SAM-Adapter can serve as an effective and adaptable tool for various downstream segmentation tasks in different fields, including medical and agriculture. This study will usher in a new era of utilizing large pre-trained image models in diverse research fields and industrial applications.

\section{Related Work}
\noindent \textbf{Semantic Segmentation.} In recent years, semantic segmentation has made significant progress, primarily due to the remarkable advancements in deep-learning-based methods such as fully convolutional networks (FCN) \cite{fcn}, encoder-decoder structures \cite{unet, bisenet, bisenetv2, segnet, sfnet}, dilated convolutions \cite{deeplabv1, deeplabv2, deeplabv3, deeplabv3+, liu2021label}, pyramid structures \cite{zhu2021learning, deeplabv3, pspnet, deeplabv3+, zhu2023continual}, attention modules \cite{acfnet, danet, ann}, and transformers \cite{zheng2021rethinking, xie2021segformer, strudel2021segmenter, cheng2022masked}. Building upon previous research, Segment Anything (SAM) \cite{SAM} introduces a large ViT-based model trained on a large visual corpus. This work aims to leverage the SAM to solve specific downstream image segmentation tasks.\\

\noindent \textbf{Adapters. }The concept of Adapters was first introduced in the NLP community \cite{houlsby2019parameter} as a tool to fine-tune a large pre-trained model for each downstream task with a compact and scalable model. In \cite{stickland2019bert}, multi-task learning was explored with a single BERT model shared among a few task-specific parameters. In the computer vision community, \cite{li2022exploring} suggested fine-tuning the ViT \cite{dosovitskiy2020image} for object detection with minimal modifications. Recently, ViT-Adapter \cite{chen2022vision} leveraged Adapters to enable a plain ViT to perform various downstream tasks. \cite{liu2023explicit} introduce an Explicit Visual Prompting (EVP) technique that can incorporate explicit visual cues to the Adapter. However, no prior work has tried to apply Adapters to leverage pretrained image segmentation model SAM trained at large image corpus. Here, we mitigate the research gap. 

\noindent \textbf{Camouflaged Object Detection (COD).} Camouflaged object detection, or concealed object detection is a challenging but useful task that identifies objects blend in with their surroundings. COD has wide applications in medicine, agriculture, and art. Initially, researches of camouflage detection relied on low-level features like texture, brightness, and color \cite{feng2013camouflage,pike2018quantifying,hou2011detection,sengottuvelan2008performance} to distinguish foreground from background. It is worth noting that some of these prior knowledge is critical in identifying the objects, and is used to guide the neural network in this paper.

Le et al.\cite{le2019anabranch} first proposed an end-to-end network consisting of a classification and a segmentation branch. Recent advances in deep learning-based methods have shown a superior ability to detect complex camouflaged objects \cite{fan2020camouflaged, mei2021camouflaged, lin2023frequency}. In this work, we leverage the advanced neural network backbone (a foundation model -- SAM) with the input of task-specific prior knowledge to achieve the state-of-the-art (SOTA) performance. 

\noindent \textbf{Shadow Detection.} Shadows can occur when an object surface is not directly exposed to light. They offer hints on light source direction and scene illumination that can aid scene comprehension \cite{karsch2011rendering,lalonde2012estimating}. They can also negatively impact the performance of computer vision tasks \cite{nadimi2004physical,cucchiara2003detecting}. Early method use hand-crafted heuristic cues like chromacity, intensity and texture \cite{huang2011characterizes,lalonde2012estimating,zhu2010learning}. Deep learning approaches leverage the knowledge learnt from data and use delicately designed neural network structure to capture the information (e.g. learned attention modules) \cite{le2018a+,cun2020towards,zhu2018bidirectional}. This work leverage the heuristic priors with large neural network models to achieve the state-of-the-art (SOTA) performance. 

\section{Method}
\subsection{Using SAM as the Backbone}
As previously illustrated, the goal of the SAM-Adapter is to leverage the knowledge learned from the SAM. Therefore, we use SAM as the backbone of the segmentation network. The image encoder of SAM is a ViT-H/16 model with 14x14 windowed attention and four equally-spaced global attention blocks. We keep the weight of pretrained image encoder frozen. 
We also leverage the mask decoder of the SAM, which consists of a modified transformer decoder block followed
by a dynamic mask prediction head. We use the pretrained SAM's weight to initialize the weight of the mask decoder of our approach and tune the mask decoder during training. We input no prompts into the original mask decoder of SAM. 

\begin{figure*}[t]
\centering
\includegraphics[width=\linewidth]{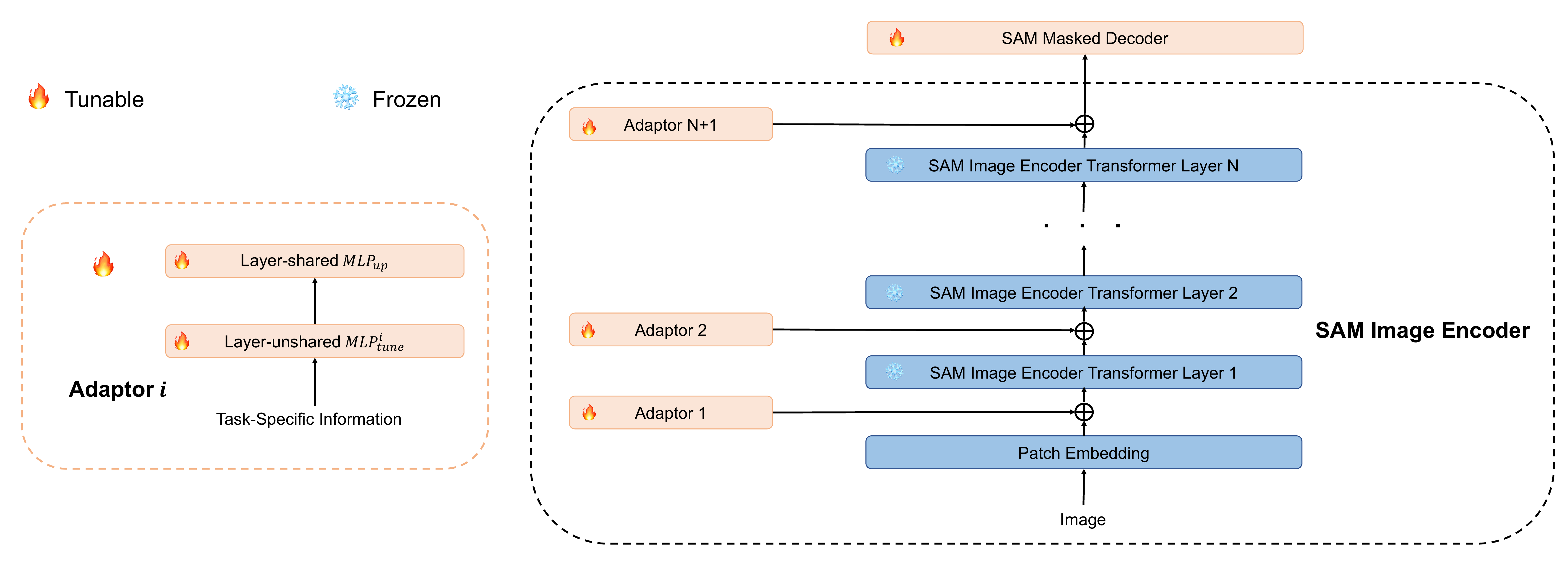}
\caption{\textbf{The architecture of the proposed SAM-Adapter.} } \label{framework}
\end{figure*}
\subsection{Adapters}
Next, the task-specific knowledge $F^i$ is learned and injected into the network via Adapters. We employ the concept of prompting, which utilizes the fact that foundation models have been trained on large-scale datasets. Using appropriate prompts to introduce task-specific knowledge \cite{liu2023explicit} can enhance the model's generalization ability on downstream tasks, especially when annotated data is scarce. 

The architecture of the proposed SAM-Adapter is illustrated in Figure \ref{framework}. We aim to keep the design of the adapter to be simple and efficient. Therefore, we choose to use an adapter that consists of only two MLPs and an activate function within two MLPs \cite{liu2023explicit}. Specifically, the adapter takes the information $F^i$ and obtains the prompt $P^i$: 
\begin{equation}
\label{get_context}
     P^i = {\rm MLP}_{up}\left({\rm GELU}\left({\rm MLP}_{tune}^i\left(F_i\right)\right)\right)
\end{equation}
in which ${\rm MLP}_{tune}^i$ are linear layers used to generate task-specific prompts for each Adapter. ${\rm MLP}_{up}$ is an up-projection layer shared across all Adapters that adjusts the dimensions of transformer features. $P^i$ refers to the output prompt that is attached to each transformer layer of SAM model. ${\rm GELU}$ is the GELU activation function \cite{hendrycks2016gaussian}. 
The information $F^i$ can be chosen to be in various forms.

\subsection{Input Task-Specific Information}
It is worth noting that the information $F^i$ can be in various forms depending on the task and flexibly designed. For example, it can be extracted from the given samples of the specific dataset of the task in some form, such as texture or frequency information, or some hand-crafted rules. Moreover, the $F^i$ can be in a composition form consisting multiple guidance information:
\begin{align}
F_i & = \sum_{1}^{N}w_jF_j 
\end{align}
in which $F^j$ can be one specific type of knowledge/features and $w^j$ is an adjustable weight to control the composed strength. 

\section{Experiments}
\subsection{Tasks and Datasets}
We select two challenging low-level structural segmentation task for SAM -- camouflaged object detection and shadow detection. For camouflaged object detection, we choose COD10K dataset \cite{fan2020camouflaged}, CHAMELEON dataset \cite{skurowski2018animal}, and CAMO dataset \cite{le2019anabranch} in our experiment. COD10K is the largest dataset for camouflaged object detection containing 3,040 training and 2,026 testing samples. CHAMELEON includes 76 images collected from the Internet for testing. CAMO dataset consists of 1250 images (1000 images for the training set and 250 images for the testing set). Following the training protocol in \cite{fan2020camouflaged}, we use combined dataset of CAMO and COD10K (the training set of camouflaged images) for training, and use the test set of CAMO, COD10K and the entire CHAMELEON dataset for performance evaluation. For shadow detection, we use ISTD dataset \cite{wang2018stacked}, which contains 1,330 training images and 540 test images. We choose kvasir-SEG \cite{jha2020kvasir} for polyp segmentation (medical image segmentation) task, and the train-test split follows the settings in Medico multimedia task at mediaeval 2020: Automatic polyp segmentation \cite{jha2020medico}. For evaluation metrics, we follow the protocol in \cite{liu2023explicit} and use commonly-used S-measure ($S_m$), mean E-measure ($E_\phi$), and MAE for evaluation of camouflaged object detection. We use balance error rate (BER) for shadow detection. We use  
For SAM, We use the official implementation and tried different prompting approaches. 

 \subsection{Implementation Details}
In the experiment, we choose two types of visual knowledge, patch embedding $F_{pe}$ and high-frequency components $F_{hfc}$, following the same setting in \cite{liu2023explicit}, which has been demonstrated effective in various of vision tasks. $w^j$ is set to 1. Therefore, the $F_i$ is derived by $F_i=F_{hfc}+F_{pe}$. 

The ${\rm MLP}_{tune}^i$ has 32 linear layers and ${\rm MLP}_{up}^i$ is one linear layer that maps the output from GELU activation to the number of inputs of the transformer layer. We use ViT-H version of SAM. Balanced BCE loss is used for shadow detection. 
BCE loss and IOU loss are used for camouflaged object detection and polyp segmentation. AdamW optimizer is used for all the experiments. The initial learning rate is set to 2e-4. Cosine decay is applied to the learning rate. The training of camouflaged object segmentation is performed for 20 epochs. Shadow segmentation is trained for 90 epochs. Polyp segmentation is trained for 120 epochs. The experiments are implemented using PyTorch on four NVIDIA Tesla A100 GPUs. 

\subsection{Experimental Result for Camouflaged Object Detection}
\begin{figure*}[h]
\centering
\includegraphics[width=\linewidth]{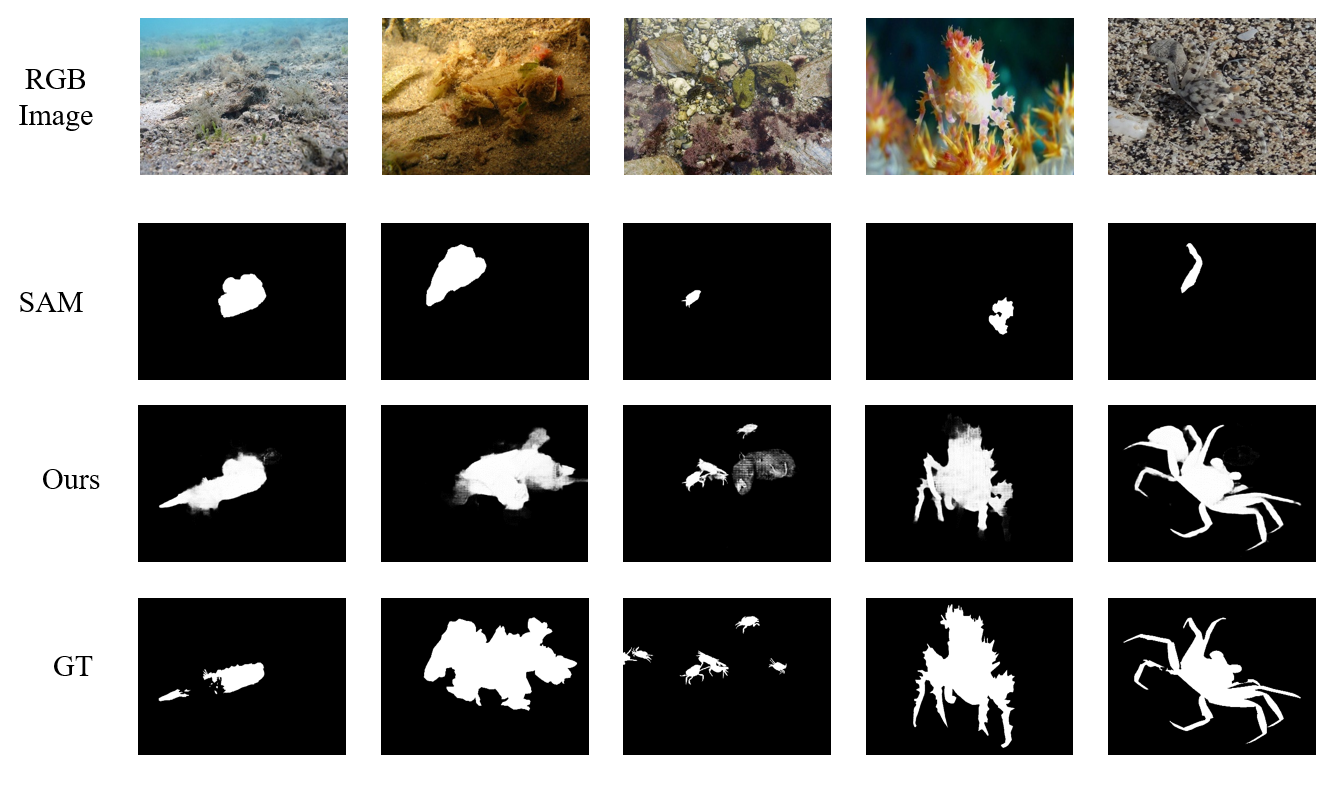}
\caption{\textbf{The Visualization Results of Camouflaged Image Segmentation.} As illustrated in the figure, the SAM failed to perceive those animals that are visually ‘hidden’/concealed in their natural surroundings. By using SAM-Adapter, our approach can significantly elevate the performance of object segmentation with SAM. The samples are from the COD-10K dataset, for other dataset, please refer to \textit{More Results} section.} \label{fig:1}
\end{figure*}

We first evaluate SAM in camouflaged object detection task, which is a very challenging task as foreground objects are often
with visual similar patterns to the background. Our experiments revealed that SAM did not perform well in this task. As shown in Figure \ref{fig:1}, SAM failed to detect some concealed objects. This can be further confirmed by the quantitative results presented in Table 1. In fact, SAM's performance was significantly lower than the existing state-of-the-art methods in all metrics evaluated. 

In Figure \ref{fig:1}, it can be found clearly that by introducing the SAM-Adapter, our method significantly elevates the performance of the model (+17.9\% in $S_\alpha$). Our approach successfully identifies concealed objects, as evidenced by clear visual results. Quantitative results also show that our method outperforms the existing state-of-the-art method. 
\\
\\
\begin{table}[t]
\scalebox{0.8}{
\begin{tabular}{c||cccc|cccc|cccc}
\hline
\multirow{2}{*}{Method} & \multicolumn{4}{c|}{CHAMELEON  \cite{skurowski2018animal}}                                    & \multicolumn{4}{c|}{CAMO \cite{le2019anabranch}}                                         & \multicolumn{4}{c}{COD10K \cite{fan2020camouflaged}}                                        \\ \cline{2-13} 
& $ S_\alpha \uparrow$               & $E_\phi \uparrow$              & $F^\omega_\beta \uparrow$              & MAE $\downarrow$           & $S_\alpha \uparrow$              & $E_\phi \uparrow$              & $F^\omega_\beta \uparrow$               & MAE $\downarrow$           & $S_\alpha \uparrow$              & $E_\phi \uparrow$              & $F^\omega_\beta \uparrow$               & MAE $\downarrow$           \\ \hline
SINet\cite{SINet}                   & 0.869          & 0.891          & 0.740          & 0.440          & 0.751          & 0.771          & 0.606          & 0.100          & 0.771          & 0.806          & 0.551          & 0.051          \\
RankNet\cite{RankNet}                 & 0.846          & 0.913          & 0.767          & 0.045          & 0.712          & 0.791          & 0.583          & 0.104          & 0.767          & 0.861          & 0.611          & 0.045          \\
JCOD \cite{JCOD}                   & 0.870          & 0.924          & -              & 0.039          & 0.792          & 0.839          & -              & 0.82           & 0.800          & 0.872          & -              & 0.041          \\
PFNet \cite{PFNet}                  & 0.882          & \textbf{0.942} & 0.810          & 0.330          & 0.782          & 0.852          & 0.695          & 0.085          & 0.800          & 0.868          & 0.660          & 0.040          \\
FBNet  \cite{FBNet}                 & 0.888          & 0.939          & \textbf{0.828} & \textbf{0.032} & 0.783          & 0.839          & 0.702          & 0.081          & 0.809          & 0.889          & 0.684          & 0.035          \\
SAM   \cite{SAM}    & 0.727 &  0.734   &  0.639    &  0.081  & 0.684 & 0.687 & 0.606 & 0.132  &  0.783    &    0.798     & 0.701 & 0.050   \\
\hline
\textbf{SAM-Adapter (Ours)}                    & \textbf{0.896} & 0.919          & 0.824          & 0.033          & \textbf{0.847} & \textbf{0.873}          & \textbf{0.765}          & \textbf{0.070}          & \textbf{0.883} & \textbf{0.918} & \textbf{0.801} & \textbf{0.025} \\ \hline
\end{tabular}}
\caption{Quantitative Result for Camouflage Detection}
\end{table}

\begin{figure*}
\centering
\includegraphics[width=0.8\linewidth]{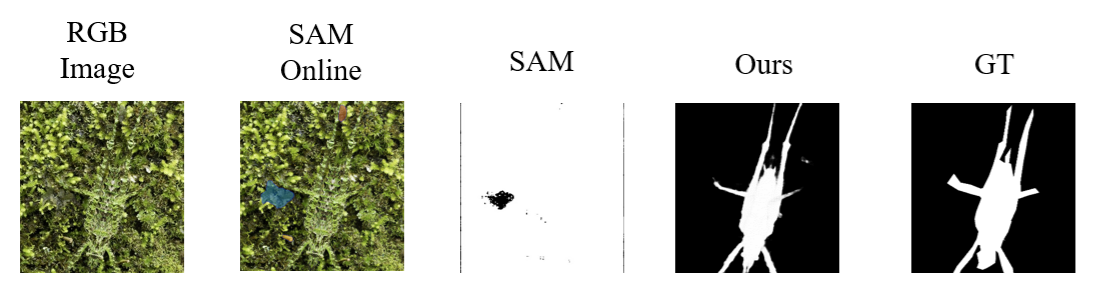}
\caption{\textbf{The Visualization Results of Camouflaged Image Segmentation with Different Prompting Approach of SAM.} The difference of this evaluation approach is that we use the SAM with input point prompts sampled in a unified manner across the image (the \textit{everything} mode that produce multiple masks of the SAM online demo, denoted \textit{SAM online} in the figure), and no input points but a mask box with the size of the image as the prompt, denoted \textit{SAM}. It can be found that in different prompting mode, SAM cannot fully identify the object. By using SAM-Adapter, our approach can significantly elevate the performance of object segmentation with SAM.} \label{fig:2}
\end{figure*}

\subsection{Experimental Result for Shadow Detection}
\begin{figure*}[h]
\centering
\includegraphics[width=0.8\linewidth]{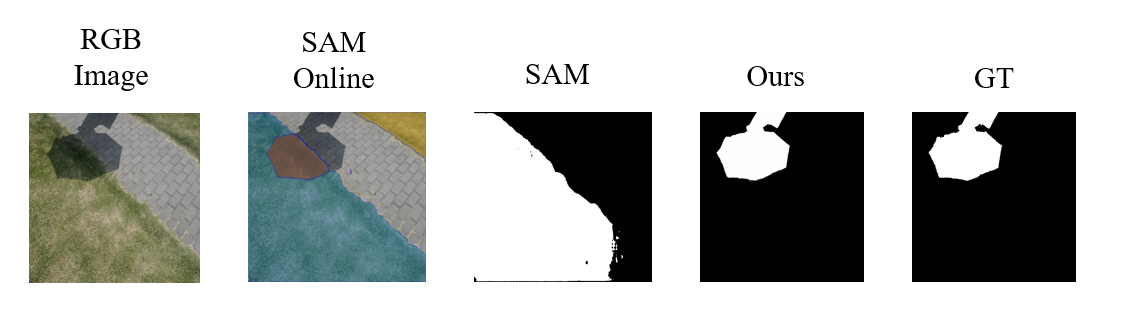}
\caption{\textbf{Shadow Detection with Different Prompting Approach of SAM.} We use SAM with input point prompts sampled in a unified manner across the
image (\textit{SAM online} in the figure), and a box of a whole image (\textit{SAM} in the figure). SAM cannot fully identify the shadow in different prompting modes. By using
SAM-Adapter, our approach elevate the performance with
SAM.} \label{fig:4}
\end{figure*}

We also evaluated SAM on the task of shadow detection. However, as depicted in Figure \ref{fig:4}, SAM struggled to differentiate between the shadow and the background information with parts missing or mistakenly added. 

\begin{wraptable}{t}{.6\textwidth}{
    \begin{adjustbox}{width=0.2\columnwidth,center}
    \begin{tabular}{l | c c c}
    \toprule
    Method & BER $\downarrow$ \\
    \midrule
    Stacked CNN \cite{vicente2016large} & 8.60\\
    \midrule
    BDRAR \cite{BDRAR} & 2.69 \\
    \midrule
    DSC \cite{DSC} & 3.42\\
    \midrule
    DSD \cite{DSD} & 2.17 \\
    \midrule
    FDRNet \cite{zhu2021mitigating} & 1.55 \\
    \midrule
    SAM \cite{SAM} & 40.51\\
    SAM-Adapter (Ours) & \textbf{1.43}\\
     \bottomrule
    \end{tabular}
    \end{adjustbox}
    \caption{Quantitative Result - Shadow Detection}
    \label{ablation_components}}
\end{wraptable}

In our study, we evaluated various methods for shadow detection and found that our results were significantly poorer than existing methods. However, by integrating the \textbf{SAM-Adapter}, we were able to significantly improve the performance of our approach. The SAM-Adapter was able to improve the detection of shadow regions, making them more clearly identifiable. Our results were verified through quantitative analysis, and Table \ref{ablation_components} demonstrates the performance boost that was brought about by the SAM-Adapter for shadow detection. 

\begin{figure*}[h]
\centering
\includegraphics[width=\linewidth]{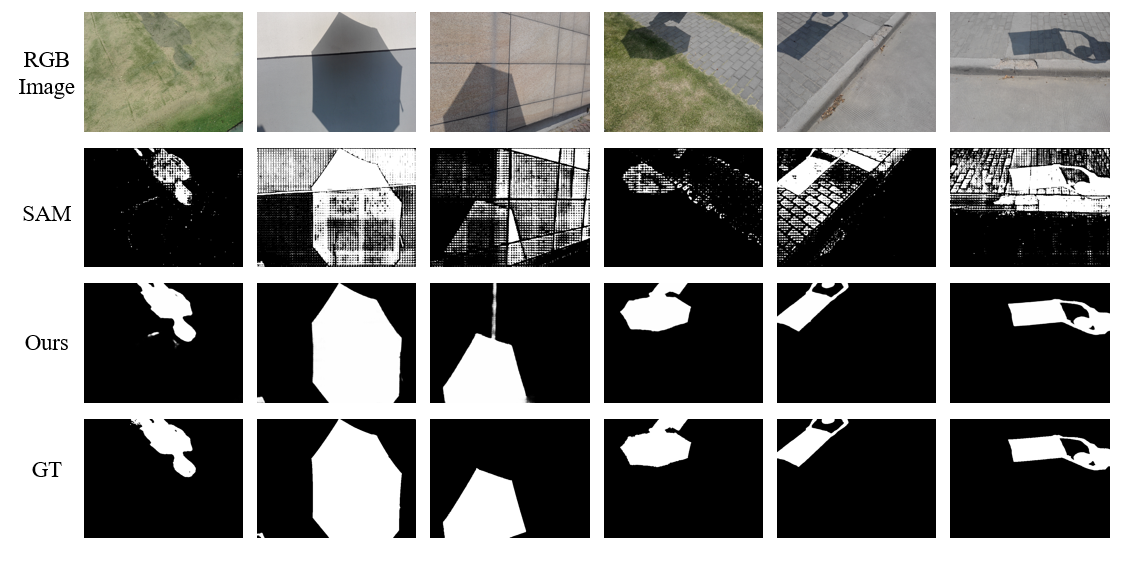}
\caption{\textbf{The Visualization Results of Shadow Detection.} As illustrated in the figure, the SAM
failed to distinguish the shadow and the background object. The \textit{SAM} is used with the box prompt with the size of a whole image as the input and no input point prompts. By using SAM-adaptor, our approach
can significantly elevate the performance of object segmentation with SAM.} \label{fig:5}
\end{figure*}

\subsection{Experimental Result for Polyp Segmentation}
We showcase an example of using SAM-Adapter in medical image segmentation. We use the example of polyp segmentation. Polyps, which can become malignant, are identified during colonoscopy and removed through polypectomy. Accurate and speedy detection and removal of polyps are critical in preventing colorectal cancer, which is a leading cause of cancer-related deaths worldwide.

Numerous deep learning approaches have been developed for identifying polyps, and while pre-trained SAM is capable of identifying some polyps, we have found that its performance can be significantly improved with our SAM-Adapter approach. The results of our study, as illustrated in Table \ref{fd} and the visualization results in Figure 6, demonstrate the effectiveness of the SAM-Adapter in enhancing the identification of polyps.

\begin{table*}{t}{
    \begin{adjustbox}{width=0.4\columnwidth,center}
    \begin{tabular}{l | c c c}
    \toprule
    Method & mDice $\uparrow$ & mIoU $\uparrow$\\
    \midrule
    UNet \cite{unet} & 0.821 & 0.756\\
    \midrule
    UNet++ \cite{zhou2018unet++} & 0.824 & 0.753 \\
    \midrule
    SFA \cite{fang2019selective} & 0.725 & 0.619 \\
    \midrule
    SAM \cite{SAM} & 0.778 & 0.707\\
    SAM-Adapter (Ours) & \textbf{0.850} & \textbf{0.776}\\
     \bottomrule
    \end{tabular}
    \end{adjustbox}
    \caption{Quantitative Result for Polyp Segmentation}
    \label{fd}}
\end{table*}

\begin{figure*}[h]
\centering
\includegraphics[width=0.9\linewidth]{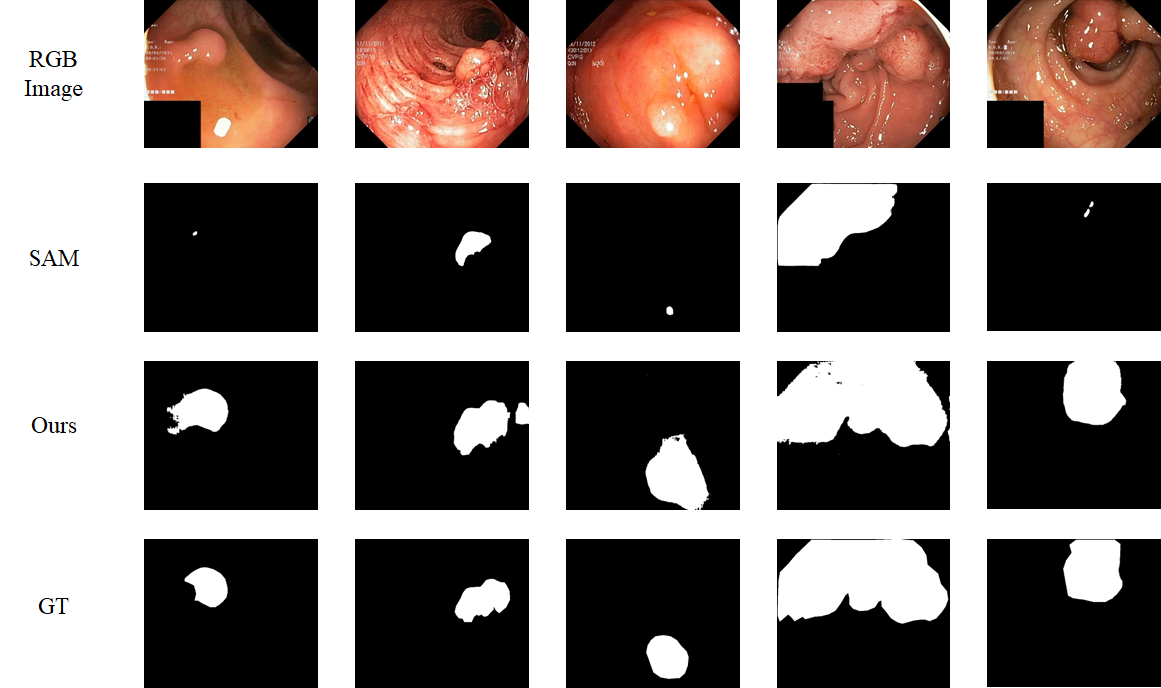}
\caption{\textbf{The Visualization Result of Polyp Segmentation.} As illustrated in the figure, although SAM can identify some polyp structures in the image, the result is not accurate. By using
SAM-Adapter, our approach elevate the performance with
SAM.} \label{fig:6}
\end{figure*}


\section{Conclusion}
In this work, we first extend the Segment Anything (SAM) model and apply it to some downstream tasks. Our experiments reveal that, like other foundational models, SAM is not effective in some vision tasks, for example, dealing with concealed objects. Therefore, we propose the SAM-Adapter, which utilizes SAM as the backbone and injects customized information into the network through simple yet effective Adapters to enhance performance in specific tasks. We evaluate our approach in camouflaged object detection and shadow detection tasks and demonstrate that the SAM-Adapter not only significantly improves SAM's performance but also achieves state-of-the-art (SOTA) results. Our approach is also capable of enhancing the performance of medical image segmentation, as we show in our polyp segmentation task. We anticipate that this work will pave the way for applying SAM in downstream tasks and will have significant impacts in various image segmentation and computer vision fields.

\section{Future Work}
This study showcases the effectiveness and versatility of using adapters and large foundation models. Moving forward, we plan to extend the SAM-Adapter to tackle even more challenging image segmentation tasks and broaden its application to other fields. We also anticipate the development of more specialized designs tailored to specific tasks. 

\bibliography{neurips_2023}
\bibliographystyle{unsrt}
\section{More Results}
\begin{figure*}[h]
\centering
\includegraphics[width=\linewidth]{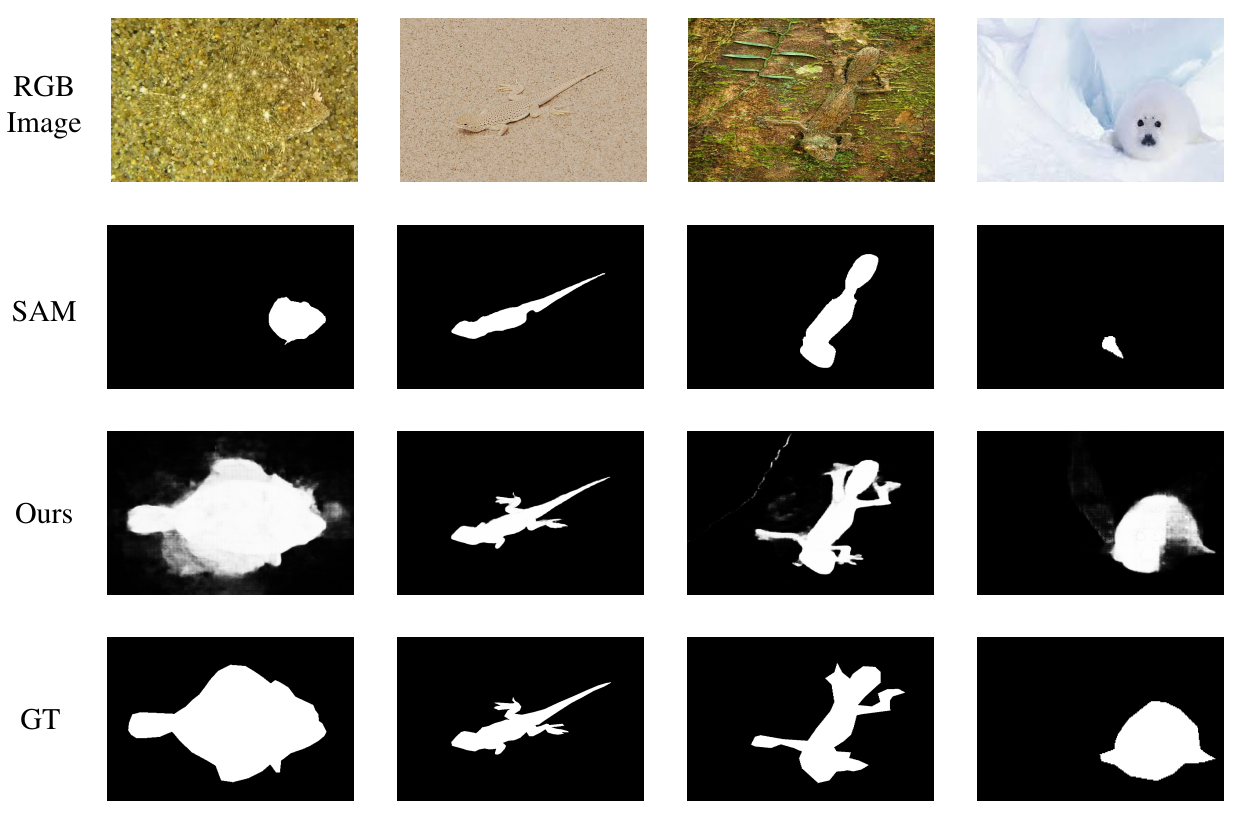}
\caption{\textbf{The Visualization Results of Camouflaged Image Segmentation of CAMO dataset.} As illustrated in the figure, the SAM failed to perceive those animals that are visually ‘hidden’/concealed in their natural surroundings. By using SAM-Adapter, our approach can significantly elevate the performance of object segmentation with SAM.} \label{fig:X}
\end{figure*}

\end{document}